\documentclass[11pt,a4paper]{article}
\usepackage[hyperref]{acl2019}
\usepackage{times}
\usepackage{latexsym}

\usepackage{url}

\usepackage{graphicx}
\usepackage{subfigure}
\usepackage{here}
\usepackage{xcolor}
\usepackage{amssymb}
\usepackage{mathtools}
\usepackage{tabularx}

\aclfinalcopy 
\def\aclpaperid{2253} 

\newcommand{\entity}[1]{\texttt{#1}}
\newcommand{\factkb}[1]{\texttt{#1}}
\newcommand{\facttext}[1]{\texttt{"#1"}}
\newcommand{\query}[1]{\emph{``#1''}}

\newcommand{\toquote}[1]{``#1''}

\title{Interpretable Question Answering on Knowledge Bases and Text}

\author{Alona Sydorova \\
  iteratec GmbH \\
  Munich, Germany \\
  \texttt{alona.sydorova@iteratec.com} \\\And
  Nina Poerner \& Benjamin Roth \\
  Center for Information and \\Language Processing \\
  LMU Munich, Germany \\
  \texttt{\{poerner,beroth\}@cis.lmu.de} \\
  }

\date{}

\begin{document}
\maketitle
\begin{abstract}
Interpretability of machine learning (ML) models becomes more relevant with their increasing adoption.
In this work, we address the interpretability of ML based question answering (QA) models on a combination of knowledge bases (KB) and text documents.
We adapt post hoc explanation methods such as LIME and input perturbation (IP) and compare them with the self-explanatory attention mechanism of the model.
For this purpose, we propose an automatic evaluation paradigm for explanation methods in the context of QA. 
We also conduct a study with human annotators to evaluate whether explanations help them identify better QA models. 
Our results suggest that IP provides better explanations than LIME or attention, according to both automatic and human evaluation. 
We obtain the same ranking of methods in both experiments, which supports the validity of our automatic evaluation paradigm.

\end{abstract}

\section{Introduction}
\label{sec:introduction}
Question answering (QA) is an important task in natural language processing and machine learning with a wide range of applications.
QA systems typically use either structured information in the form of knowledge bases (KBs), or raw text.
Recent systems have successfully combined both types of knowledge \cite{TextKBQA}.

Nowadays, due to the changing legal situation and growing application in critical domains, ML based systems are increasingly required to provide explanations of their output.
\newcite{LiptonMythos}, \newcite{ManipulationMeasuringInterpretability} and \newcite{RigorousInterpretableML} point out that there is no complete agreement on the definition, measurability and evaluation of interpretability in ML models.
Nevertheless, a number of explanation methods have been proposed in the recent literature, with the aim of making ML models more transparent for humans.

To the best of our knowledge, the problem of explanations for deep learning based QA models working on a combination of structured and unstructured data has not yet been researched.
Also, there is a lack of evaluation paradigms to compare different explanation methods in the context of QA.

\subsection*{Contributions}
\begin{list}{-}{}
\item We explore interpretability in the context of QA on a combination of KB and text. In particular, we apply attention, LIME and input perturbation (IP).
\item In order to compare these methods, we propose a novel automatic evaluation scheme based on ``fake facts''.
\item We evaluate whether explanations help humans identify the better out of two QA models.
\item We show that the results of automatic and human evaluation agree.
\item Our results suggest that IP performs better than attention and LIME in this context.
\end{list}

%

\section{Question Answering on Knowledge Bases and Text}
\label{sec:textkbqa}

\begin{figure*}
\begin{center}
\includegraphics[width=14cm]{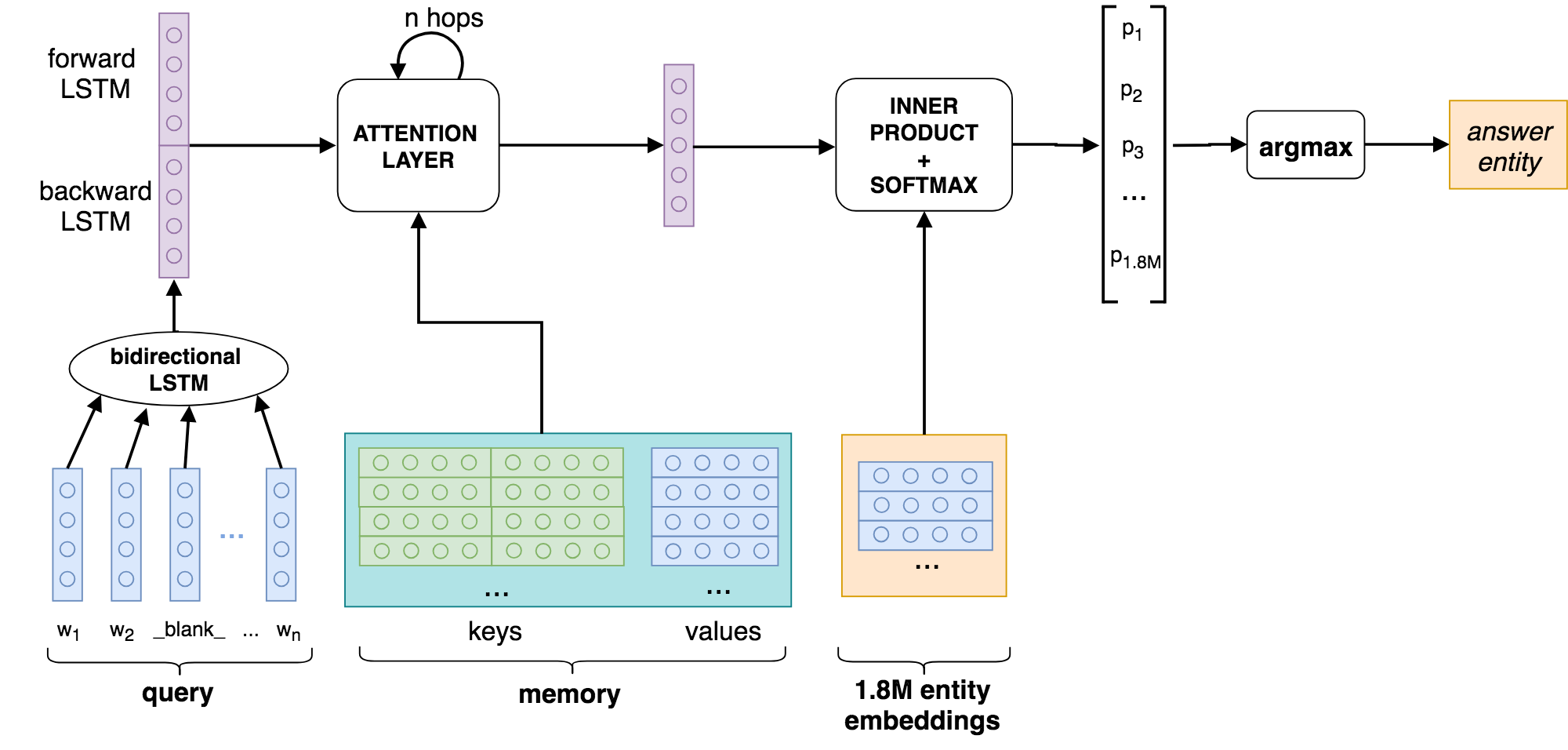}
\caption{Overview of the TextKBQA model architecture.}
\label{fig:model_architecture}
\end{center}
\end{figure*}

The combination of knowledge bases and text data is of particular interest in the context of QA.
While knowledge bases provide a collection of facts with a rigid structure, the semantic information contained in text documents has the potential to enrich the knowledge base.
In order to exploit different information sources within one QA system, \citet{TextKBQA} introduce the TextKBQA model, which works on a universal schema representation \cite{UnivSchema} of a KB and text documents. 
They state that \toquote{individual data sources help fill the weakness of the other, thereby improving overall performance} and conclude that \toquote{the amalgam of both text and KB is superior than KB alone.}
Their model solves the so-called \textit{cloze questions} task, i.e., filling in blanks in sentences. 
For example, the answer to \query{Chicago is the third most populous city in \_blank\_.} would be the entity \emph{the USA}. 
The model has a KB and a number of raw text sentences at its disposal. 
\newcite{TextKBQA} use Freebase \cite{Freebase} as KB ($8.0$M facts) and ClueWeb \cite{ClueWeb} as raw text source ($0.3$M sentences).
They test on question-answer pairs from SPADES \cite{SPADES} ($93$K queries).

The TextKBQA model (Figure \ref{fig:model_architecture}) is a key-value memory network that uses distributed representations for all entities, relations, textual facts and input questions.
Every memory cell corresponds to one KB fact or one textual fact, which are encoded as key-value pairs \cite{KeyValueMemoryNetworks}.

Every \textbf{KB fact} is a triple consisting of a subject $s$, an object $o$ and the relation $r$ between these entities.
$s,r,o$ are embedded into real-valued vectors $\mathbf{s},\mathbf{r},\mathbf{o}$.
The memory key is the concatenation of subject and relation embedding: $\mathbf{k} = [\mathbf{s}; \mathbf{r}] \in \mathbb{R}^{2d}$.
The memory value is the embedding of the object: $\mathbf{v}=\mathbf{o} \in \mathbb{R}^d$.
\textbf{Textual facts} are sentences that contain at least two entities.
They are also represented as triples, where the relation is a token sequence: $(s, [w_{1},...,arg1,...,arg2,...,w_{n}], o)$.
To convert the sentence into a vector, $arg1$ and $arg2$ are replaced by $s$ and $\_blank\_$ respectively.
Then, the sequence is processed by a bidirectional LSTM.
Its last states are concatenated to form the memory key $\mathbf{k} = [\overrightarrow{LSTM}([w_1,...,w_n]); \overleftarrow{LSTM}([w_1,...,w_n])] \in \mathbb{R}^{2d}$. 
The memory value is $\textbf{v}=\textbf{o}$, as before.

A question $q = [w_{1},...,e,...,\_blank\_,...,w_{n}]$ is transformed into a distributed representation $\mathbf{q} \in \mathbb{R}^{2d}$ using the same bidirectional LSTM as before.
In this way, KB and textual facts as well as queries are in the same $\mathbb{R}^{2d}$ space.

Given $\mathbf{q}$ and a set of relevant facts, represented by key-value pairs $(\mathbf{k}, \mathbf{v})$, TextKBQA performs multi-hop attention.
More specifically, the context vector $\mathbf{c}_0$ is set to $\mathbf{q}$.
In every iteration (\emph{hop}) $t$, a new context vector $\mathbf{c}_{t}$ is computed as:

\begin{equation}
\mathbf{c}_{t} = W_{t} \big(\mathbf{c}_{t-1} + W_{p} \sum_{(k,v)\in\mathcal{M}} \text{softmax}(\mathbf{c}_{t-1} \cdot \textbf{k})\textbf{v}\big)
\end{equation}
where $W_{p}$, $W_{t}$ are weight matrices.
In practice, $\mathcal{M}$ contains only facts that share an entity with the query.
The result of the last hop is fed into a fully-connected layer to produce a vector $\mathbf{b} \in \mathbb{R}^d$.
Then, the inner product between $\mathbf{b}$ and all entity embeddings is taken.
The entity with the highest inner product is chosen as the model's answer $a_{q}$. 

We train the TextKBQA model using the datasets described above. 
We limit the number of textual facts per query to $500$, since only $35$ out of $1.8$M entities in the dataset have more than $500$ textual facts. 
Apart from this modification, we use the exact same implementation and training setup as in \newcite{TextKBQA}. 
Our final model achieves an $F_1$ score of $41.59$ on the dev dataset and $40.27$ on the test dataset, which is slightly better than the original paper ($41.1$ and $39.9$, respectively).

\section{Explanation methods}
\label{sec:expl_methods}

We first present some important notation and give a working definition of an explanation method. 

Formally, let $\mathbb{F}$  be a database consisting of all KB and textual facts: $\mathbb{F} = \mathbb{F}_{KB} \cup \mathbb{F}_{text}$. 
Furthermore, let $\mathcal{E}$ be a set of \textbf{entities} that are objects and subjects in $\mathbb{F}$, and let $\mathcal{R}$ be a set of \textbf{relations} from $\mathbb{F}_{KB}$.
In the following we will use a general notation $f$ for a fact from $\mathbb{F}$, distinguishing between KB and textual facts only when necessary.
 
Let $q$ be a \textbf{query}, and $\mathcal{F}\subseteq \mathbb{F}$ the corresponding set of facts, such that for $\forall f\in\mathcal{F}$ holds: $subject_{f} \in q$. 
Let $TextKBQA$ be a function computed by the TextKBQA model and $a_{q} = TextKBQA(q, \mathcal{F})$, $a_{q} \in \mathcal{E}$, the predicted \textbf{answer} to the query $q$.
Note that $a_{q}$ is not necessarily the ground truth answer for $q$.

Analogously to \newcite{hybrid_document}, we give the following definition: 
an \textit{explanation method} is a function $\phi(f, a_{q}, q, \mathcal{F})$ that assigns real-valued relevance scores to facts $f$ from $\mathcal{F}$ given an input query $q$ and a target entity $a_{q}$.
If $\phi(f_{1}, a_{q}, q, \mathcal{F}) > \phi(f_{2}, a_{q}, q, \mathcal{F})$ then fact $f_{1}$ is of a higher relevance for $a_{q}$ given $q$ and $\mathcal{F}$ than fact $f_{2}$.


\subsection{Attention Weights}
\label{sec:attn_weights}
The attention mechanism provides an explanation method which is an integral part of the TextKBQA architecture.

We formally define the explanation method \textbf{attention weights} as: 

\begin{equation}
\phi_{aw}(f, a_{q}, q, \mathcal{F}) = \text{softmax}(K_{\mathcal{F}} \cdot \mathbf{q})_{f}
\end{equation}
where $K_{\mathcal{F}}$ is a matrix whose rows are key vectors of facts in $\mathcal{F}$.

Since the TextKBQA model takes three attention hops per query, $\phi_{aw}$ can be extended as follows:
On the one hand, we can take attention weights from the first, second or third (=last) hops.
Intuitively, attention weights from the first hop reflect the similarity of fact keys with the original query, while attention weights from the last hop reflect the similarity of fact keys with the summarized context from all previous iterations.
On the other hand, some aggregation of attention weights could also be a plausible explanation method. 
For every fact, we take the mean attention weight over hops to be its average relevance in the reasoning process.

Taking into account the above considerations we redefine $\phi_{aw}$:
\begin{itemize}
\item[$-$] \textbf{attention weights at hop $j$}:
{
\small
\begin{equation}
\phi_{aw_{j}}(f, a_{q}, q, \mathcal{F}) = \text{softmax}(K_{\mathcal{F}} \cdot \mathbf{c}_{j-1})_{f}
\end{equation}
}
\item[$-$] \textbf{average attention weights}: 
{
\small
\begin{equation}
\phi_{aw_{avg}}(f, a_{q}, q, \mathcal{F}) = 
\frac{1}{h} \sum_{j=1}^{h} \text{softmax}(K_{\mathcal{F}} \cdot \mathbf{c}_{j-1})_{f}
\end{equation}
}
 where $h$ is the number of hops.
\end{itemize}

\subsection{LIME}
\label{subsec:lime_textkbqa}
LIME (Local Interpretable Model-Agnostic Explanations) is a model-agnostic explanation method \cite{LIME}. 
It approximates behavior of the model in the vicinity of an input sample with the help of a less complex, interpretable model.   

LIME requires a mapping from original features (used by TextKBQA) to an interpretable representation (used by LIME). 
For this purpose we use binary ``bag of facts'' vectors, analogously to the idea of bag of words: a vector $z\in\{ 0,1 \}^{|\mathcal{F}|}$ indicates presence or absence of a fact $f$ from $\mathcal{F}$.
The reverse mapping is straightforward.

We first turn the original fact set $\mathcal{F}$ into an interpretable representation $z$. 
Every entry of this vector represents a fact from $\mathcal{F}$. 
Then we sample vectors $z'$ of the same length $|\mathcal{F}|$ by drawing facts from $\mathcal{F}$ using the Bernoulli distribution with $p=0.5$.
In every $z'$ vector, the presence or absence of facts is encoded as 1 or 0, respectively.
We set the number of samples to 1000 in our experiments.

For every $z'$, we obtain the corresponding original representation $\mathcal{F'}$ and give this reduced input to the TextKBQA model.
Note that the query $q$ remains unchanged. 
We are interested in the probability that $a_{q}$ is still the predicted answer to the query $q$, given facts $\mathcal{F'}$ instead of $\mathcal{F}$. 
In the TextKBQA model, this probability is obtained from the inner product of $\mathbf{b}$ and the entity embedding matrix $E$ at position $a_{q}$. 
We define this step as a function $logit(q, \mathcal{F},  a_{q}) = (E \cdot \mathbf{b})_{a_{q}}$.

We gather the outputs of $logit(q, \mathcal{F'}, a_{q})$ for all sampled instances, together with the corresponding binary vectors, into a dataset $\mathcal{Z}$.
Then, we train a linear model on $\mathcal{Z}$ by optimizing the following equation: 
\begin{equation}
\xi (q, \mathcal{F}) = \mathop{\mathrm{argmin}}_{g\in G} \mathcal{L} (logit, g) 
\end{equation}
where $\mathcal{L}$ is ordinary least squares and $G$ is the class of linear models, such that $g(z') = w_{g}\cdot z'$.\footnote{We do not use a proximity measure, because, unlike the original LIME, we only sample from the facts currently present in $\mathcal{F}$, and not from the whole $\mathbb{F}$ set.}

From the linear model $g$, we extract a weight vector $w_{g} \in \mathbb{R}^{|\mathcal{F}|}$.
This vector contains LIME relevance scores for facts in $\mathcal{F}$ given $a_{q}$ and $q$. 
We formally define the \textbf{LIME} explanation method for the TextKBQA model as follows:
\begin{equation}
\phi_{lime}(f, a_{q}, q, \mathcal{F}) = w_{g, f}
\end{equation}

\subsection{Input Perturbation Method}
Another explanation method is input perturbation (IP), originally proposed by \citet{occlusion_based_ip}, who apply it on a sentiment analysis task. 
They compute relevance scores for every word in a dictionary as the average relative log-likelihood difference that arises when the word is replaced with a baseline value.
This method cannot be directly applied to QA, because the same fact can be highly relevant for one query and irrelevant for another.
Therefore, we constrain the computation of log-likelihood differences to a single data sample (i.e., a single query).

We formally define the \textbf{input perturbation} (IP) explanation method as follows:

\small
\begin{equation}
\phi_{ip}(f, a_{q}, q, \mathcal{F}) = \frac{logit(q, \mathcal{F},  a_{q}) - logit(q, \mathcal{F}\setminus\{f\},  a_{q})}{logit(q, \mathcal{F},  a_{q})} 
\end{equation}

\normalsize
where $logit$ is the same logit function that we used for LIME. 
A positive difference means that if we remove fact $f$ when processing query $q$, the model's hidden vector $\mathbf{b}$ is less similar to the entity $a_{q}$, suggesting that the fact is relevant.

%

\section{Automatic evaluation using fake facts}
\label{sec:automatic_evalution}
This section presents our automatic evaluation approach, which is an extension of the hybrid document paradigm \cite{hybrid_document}. 
The major advantage of automatic evaluation in the context of explanation methods is that it does not require manual annotation.

\subsection{Definition of automatic evaluation}
\citet{hybrid_document} create hybrid documents by randomly concatenating fragments of different documents.
We adapt this paradigm to our use case in the following way:

Let $q$ be a query and $\mathcal{F}$ the corresponding set of facts.
We define the corresponding \textbf{hybrid fact set} $\hat{\mathcal{F}}$ as the union of $\mathcal{F}$ with another disjoint fact set $\mathcal{F'}$: 
\begin{equation}
\hat{\mathcal{F}}=\mathcal{F}\cup\mathcal{F'}\text{, where } \mathcal{F}\cap\mathcal{F'}=\emptyset.
\end{equation}

Conceptually, $\mathcal{F'}$ are ``fake facts''.
We discuss how they are created below; for now, just assume that TextKBQA is unable to correctly answer $q$ using only $\mathcal{F'}$.
Note that we only consider queries that are correctly answered by the model based on their hybrid fact set $\hat{\mathcal{F}} = \mathcal{F} \cup \mathcal{F'}$.

The next step is to obtain predictions $a_q$ for the hybrid instances and to explain them with the help of an explanation method $\phi$.
Recall that $\phi$ produces one relevance score per fact.
The fact with the highest relevance score, $rmax(\hat{\mathcal{F}}, q, \phi)$, is taken to be the most relevant fact given query $q$, answer $a_q$ and facts $\hat{\mathcal{F}}$, according to $\phi$.
We assume that $\phi$ made a reasonable choice if $rmax(\hat{\mathcal{F}}, q, \phi)$ stems from the original fact set $\mathcal{F}$ and not from the set of fake facts $\mathcal{F'}$.

Formally, a ``hit point'' is assigned to $\phi$ if:

\begin{equation}
  hit(\phi, q, \hat{\mathcal{F}})=\begin{cases}
    1, & \text{if $rmax(\hat{\mathcal{F}}, q, \phi) \in \mathcal{F}$},\\
    0, & \text{if $rmax(\hat{\mathcal{F}}, q, \phi) \in \mathcal{F'}$}.
  \end{cases}
\end{equation}

The \textbf{pointing game accuracy} of explanation method $\phi$ is simply its number of hit points divided by the maximally possible number of hit points.




\begin{table*}
\centering
 	\begin{tabular}{ l | l }
 & \textbf{Facts in hybrid fact set $\hat{\mathcal{F}}$} \\
\hline
\textbf{Facts from $\mathcal{F}$} & Disney award.award\_honor.award\_winner\_award.award\_honor.honored\_for Bambi \\
(real facts) & Disney's Steamboat Willie premiered on November 18th 1928 at the Broadway.\\
 & \color{green} \underline{Disney film.performance.actor\_film.performance.character Mickey} \\
 & Disney film.film.directed\_by.2\_film.film.directed\_by.1 The Opry House \\
\hline
\textbf{Facts from $\mathcal{F'}$}  & But in the summer of 2007, Apple rocked Disney by launching the iPhone.  \\
(fake facts) & \color{red} \textit{Disney fashion.clothing\_size.region\_fashion.clothing\_size.person Frankie Rayder} \\
 & The Libertarian is a Disney political party created in 1971.\\
 & eBay is the largest marketplace in the Disney. \\
  \end{tabular}
  
 \caption{An example of a hybrid instance. Query: \query{Walt Disney himself was the original voice of  \_blank\_.}. Answer: \entity{Mickey}. 
 Green underlined: fact with the maximal relevance score assigned by IP.
 Red italics: fact with the maximal relevance score assigned by average attention weights.}
\label{tab:hit_strict_example}
\end{table*}

\subsection{Creating Fake Facts}
To create fake facts for query $q$, we randomly sample a different query $q'$ that has the same number of entities and gather its fact set $\mathcal{F'}$.
We then replace subject entities in facts from $\mathcal{F'}$ with subject entities from $\mathcal{F}$.  
We call these ``fake facts'' because they do not exist in $\mathbb{F}$, unless by coincidence.
For example, let $q$ be \query{\_blank\_ was chosen to portray \underline{Patrick Bateman}, a \underline{Wall Street} serial killer.} and $q'$ be \query{This year \underline{Philip} and \_blank\_ divided \underline{Judea} into four kingdoms.}
Then replace subject entities \entity{Philip} and \entity{Judea} in facts of $\mathcal{F'}$ by subject entities \entity{Patrick Bateman} and  \entity{Wall Street}, respectively. 
E.g., the KB fact \factkb{Philip people.person.gender.1 Males} is turned into \factkb{Patrick Bateman people.person.gender.1 Males}, the textual fact \facttext{This year Herod divided Judea into four kingdoms.} becomes \facttext{This year Herod divided Wall Street into four kingdoms.}

Our assumption is as follows: If the model is still able to predict the correct answer despite these fake facts, then this should be due to a fact contained in $\mathcal{F}$ and not in $\mathcal{F}'$.
This assumption fails when we accidentally sample a fact that supports the correct answer.
Therefore, we validate $\mathcal{F'}$ by testing whether the model is able to predict the correct answer to $q$ using just $\mathcal{F'}$.
If this is the case, a different query $q'$ and a different fake fact set $\mathcal{F'}$ are sampled and the validation step is applied again. 
This procedure goes on until a valid $\mathcal{F'}$ is found.

Table~\ref{tab:hit_strict_example} contains an example of a query with real and fake facts for which explanations were obtained by average attention weights and IP.
IP assigns maximal relevance to a real fact from $\mathcal{F}$, which means that $\phi_{ip}$ receives one hit point for this instance.
The average attention weight method considers a fake fact from $\mathcal{F'}$ to be the most important fact and thus does not get a hit point.

\subsection{Experiments and results}
We perform the automatic evaluation experiment on the test set, which contains 9309 question-answer pairs in total.
Recall that we discard queries that cannot be answered correctly, which leaves us with 2661 question-answer pairs.
We evaluate the following explanation methods:
\begin{itemize}
 \item $\phi_{aw_{1}}$ - attention weights at first hop
 \item $\phi_{aw_{3}}$ - attention weights at third (last) hop
 \item $\phi_{aw_{avg}}$ - average attention weights
 \item $\phi_{lime}$ - LIME with 1000 samples per instance
 \item $\phi_{ip}$ - input perturbation (IP)
 \end{itemize} 
 
A baseline that samples a random fact for $rmax(...)$ is used for reference.

Table~\ref{tab:hitscore_strict} shows pointing game accuracies and the absolute number of hit points achieved by all five explanation methods and the baseline.
All methods beat the random baseline.

 \textbf{IP} is the most successful explanation method with a pointing game accuracy of $0.97$, and \textbf{LIME} comes second.
Note that we did not tune the number of samples per query drawn by LIME, but set it to $1000$.
It is possible that as a consequence, queries with large fact sets are not sufficiently explored by LIME.
On the other hand, a high number of samples is computationally prohibitive, as TextKBQA has to perform one inference step per sample.


\textbf{Attention weights at hop 3} performs best among the attention-based methods, but worse than LIME and IP. We suspect that the last hop is especially relevant for selecting the answer entity.
The poor performance of attention is in line with recent work by \newcite{attentionNotExplanation}, who also question the validity of attention as an explanation method.

We perform significance tests by means of binomial tests (with $\alpha =0.05$). 
Our null hypothesis is that there is no significant difference in hit scores between a given method and the next-highest method in the ranking in Table~\ref{tab:hitscore_strict}.
Differences are statistically significant in all cases, except for the difference between attention weights at hop 3 and average attention weights ($p=0.06$).

 \begin{table*}

 	\centering
 \begin{tabular}{c||c|c}
 
 \textbf{Explanation method} & \textbf{Hit points} & \textbf{Pointing game acc.} \\ 
 
\hline
attention weights at hop 1 & 1849 & 0.69 \\
\hline
attention weights at hop 3 & 2116 & 0.80 \\
\hline
average attention weights & 2081 & 0.78 \\
\hline
LIME & 2271 & 0.85 \\
\hline
IP & 2570 & 0.97 \\
\hline
\hline
random baseline & 1458 & 0.55

  \end{tabular}
 \caption{Hit points and pointing game accuracy. 2661 out of 9309 test set questions were used.}
 \label{tab:hitscore_strict}
  \end{table*}

\section{Evaluation with human annotators}
\label{sec:human_study}

The main goal of explanation methods is to make machine learning models more transparent for humans.
That is why we conduct a study with human annotators.

Our experiment is based on the trust evaluation study conducted by \newcite{trustWorthinessStudy} which, in turn, is motivated by the following idea: An important goal of interpretability is increasing users' trust in ML models, and trust is directly impacted by how much a model is understood \cite{LIME}. 
\newcite{trustWorthinessStudy} develop a method to visualize explanations for convolutional neural networks on an image classification task, and evaluate this method in different ways.

One of their experiments is conducted as follows:
Given two models, one of which is known to be better (e.g., to have higher accuracy), instances are chosen that are classified correctly by both models. 
Visual explanations for these instances are produced by the evaluated explanation methods, and human annotators are given the task of rating the reliability of the models relative to each other, based on the predicted label and the visualizations.
Since the annotators see only instances where the classifiers agree, judgments are based purely on the visualizations. 
An explanation method is assumed to be successful if it helps annotators identify the better model.
The study confirmed that humans are able to identify the better classifier with the help of good explanations.
 
We perform a similar study for our use case, but modify it as described below.

\subsection{Experimental setup}
We use two TextKBQA Models, which are trained differently:
\begin{itemize}
\item \textit{model A} is the model used above, with a test set $F_{1}$ of $40$
\item \textit{model B} is a TextKBQA model with a test set $F_{1}$ of $23$. The lower score was obtained by training the model for fewer epochs and without pre-training in \textsc{OnlyKB} mode (see \cite{TextKBQA}).
\end{itemize}

We only present annotators with query instances for which both models output the same answer.
However, we do not restrict these answers to be the ground truth.
We perform the study with three explanation methods: average attention weights, LIME and IP. 
We apply each of them to the same question-answer pairs, so that the explanation methods are equally distributed among tasks.

\begin{figure*}
\begin{center}
\fbox{\includegraphics[width=.6\textwidth]{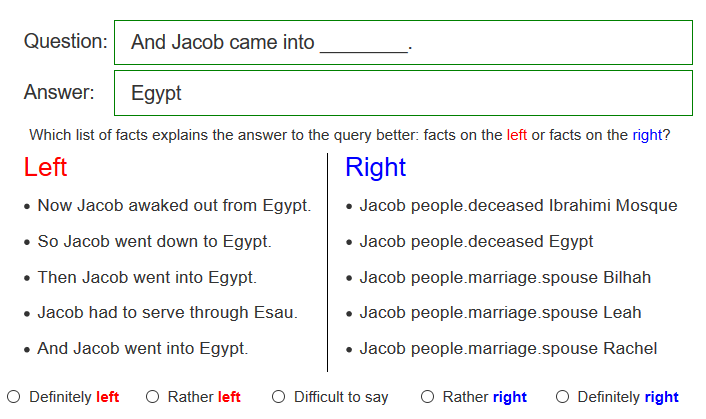}}
\end{center}
\caption{Interface for the human annotation study.}
\label{fig:human_interface}
\end{figure*}

Every task contains one query and its predicted answer (which is the same for both models), and explanations for both models by the same explanation method.
In contrast to image classification, it would not be human-friendly to show participants all input components (i.e., all facts), since their number can be up to $5500$.
Hence, we show the top5 facts with the highest relevance score.
The order in which model A and model B appear on the screen (i.e., which is ``left'' and which is ``right'' in Figure \ref{fig:human_interface}) is random to avoid biasing annotators.

Annotators are asked to compare both lists of top5 facts and decide which of them explains the answer better. 
This decision is not binary, but five options are given: \emph{definitely left}, \emph{rather left}, \emph{difficult to say}, \emph{rather right} and \emph{definitely right}.
The interface is presented in Figure \ref{fig:human_interface}.

$25$ computer science students, researchers and IT professionals took part in our study and annotated 600 tasks in total. 

\subsection{Results}
As shown in Table~\ref{tab:humaneval_results_perc}, the answer \textit{difficult to say} is the most frequent one for all explanation methods.
For attention weights and LIME there is a clear trend that, against expectations, users found fact lists coming from model B to be a better explanation.
The total share of votes for \emph{definitely model B} and \emph{rather model B} makes up $49.5\%$ for attention weights and $29\%$ for LIME, while \emph{definitely model A} and \emph{rather model A} gain $19.5\%$ and $23.5\%$, respectively.
In contrast to that, IP achieves a higher share of votes for model A than for model B: $16.5\%$ vs.\ $10.5\%$.


\begin{table*}
 	\centering
 \begin{tabular}{c||c|c|c}
 & \textbf{avg. attention weights}& \textbf{LIME} & \textbf{IP} \\ 
 \hline 
 definitely model A & $6.0\%$ & $6.5\%$ & $5.0\%$ \\ 
 \hline 
 rather model A & $13.5\%$ & $17.0\%$  & $11.5\%$ \\ 
 \hline 
  difficult to say & $31.0\%$ & $47.5\%$  & $73.0\%$ \\ 
 \hline 
  rather model B & $28.0\%$ & $18.5\%$  & $9.0\%$ \\ 
 \hline 
  definitely model B & $21.5\%$ & $10.5\%$  & $1.5\%$ \\ 
 \hline  
 \hline 
   aggregate score & $0.386$ & $0.476$  & $0.524$ \\ 
 \hline   
  \end{tabular}
 \caption{Percentage distribution of votes, and aggregate score, from the human annotation study.}
 \label{tab:humaneval_results_perc}
  \end{table*}
  
Analogously to \newcite{trustWorthinessStudy}, we compute an aggregate score that expresses how much an explanation method helps users to identify the better model.
Votes are weighted in the following way: \emph{definitely model A} $+1$, \emph{definitely model A} $+0.75$, \emph{difficult to say} $+0.5$, \emph{rather model B} $+0.25$ and \emph{definitely model B} $+0$.
We then compute a weighted average of votes for all tasks per explanation method.
In this way, scores are bounded in $[0,1]$ like the values of the hit score function used for the automatic evaluation.
Values smaller than $0.5$ indicate that the less accurate model B was trusted more, while values larger than $0.5$ represent a higher level of trust in the more accurate model A.
According to this schema, attention weights achieve a score of $0.386$ and LIME achieves a score of $0.476$.
The score of the IP method is $0.524$, which means that participants were able to identify the better model A when explanations were given by IP.

Significance tests show that while attention weights perform significantly worse than other methods, the difference between LIME and IP is insignificant, with $p=0.07$.
A larger sample of data and/or more human participants may be necessary in this case.

We also collected feedback from participants and performed qualitative analysis on the evaluated fact lists. 
The preference for the \textit{difficult to say} option can be explained by the fact that in many cases, both models were explained with the same or very similar fact lists.
In particular, we found that IP provided identical top five fact lists in $120$ out of $200$ tasks. 
In the case of attention weights and LIME, this occurs only in $9$ and $10$ cases out of $200$ tasks.

Another problem mentioned by annotators was that KB facts are not intuitive or easy to read for humans that have not dealt with such representations before.
It would be interesting to explore if some additional preprocessing of facts would lead to different results.
For example, KB facts could be converted into natural language sentences, while textual facts could be presented with additional context like the previous and the next sentences from the original document. 
We leave such preprocessing to future work.

\section{Related work}
\label{sec:related_work}

\citet{lime_query} estimate relevance of words in queries with LIME to test the robustness of QA models. However, they do not analyze the importance of the facts used by these QA systems.

\citet{quint} present a QA system called QUINT that provides a visualization of how a natural language query is transformed into formal language and how the answer is derived. 
However, this system works only with knowledge bases and the explanatory system is its integral part, i.e., it cannot be reused for other models.
\citet{IRN} propose an out-of-the-box interpretable QA model that is able to answer multi-relation questions.
This model is explicitly designed to work only with KBs.
Another approach for interpretable QA with multi-hop reasoning on knowledge bases is introduced by \citet{comp_attn_networks}. 
They claim that the transparent nature of attention distributions across reasoning steps allows humans to understand the model's behavior.

To the best of our knowledge, the interpretability of QA models that combine structured and unstructured data has not been addressed yet.
Even in the context of KB-only QA models, no comprehensive evaluation of different explanation methods has been performed.
The above-mentioned approaches also lack empirical evaluation with human annotators, to estimate how useful the explanations are to non-experts.

\section{Conclusions}
\label{sec:conclusions}
We performed the first evaluation of different explanation methods for a QA model working on a combination of KB and text.
The evaluated methods are attention, LIME and input perturbation.
To compare their performance, we introduced an automatic evaluation paradigm with fake facts, which does not require manual annotations.
We validated the ranking obtained with this paradigm through an experiment with human participants, where we observed the same ranking. 
Based on the outcomes of our experiments, we recommend the IP method for the TextKBQA model, rather than the model's self-explanatory attention mechanism or LIME.

\bibliographystyle{acl_natbib}
\bibliography{bibliography/dl,bibliography/explanation,bibliography/other,bibliography/textkbqa,bibliography/related_work}

\end{document}